\documentclass[10pt,conference]{IEEEtran}
\IEEEoverridecommandlockouts
\usepackage{float}
\usepackage{tabularx}
\usepackage{booktabs}
\usepackage{tabularx}
\usepackage{multirow}
\usepackage{makecell}
\usepackage{graphicx}
\usepackage{subcaption} 
\usepackage{cite}
\usepackage{amsmath,amssymb,amsfonts}
\usepackage{algorithmic}
\usepackage{graphicx}
\usepackage{textcomp}
\usepackage{xcolor}
\usepackage[pagebackref,breaklinks,colorlinks]{hyperref}
\usepackage{multirow}

\begin{document}

\title{I2PRef: Image-Driven Point Completion with Iterative Refinement\\
\thanks{This work was funded by the Bavarian Ministry of Economic Affairs, Regional Development and Energy (project BAVAR-RADAR, label DIK0622), and partly supported by Deutsche Forschungsgemeinschaft (DFG): Projects 532402151 and SFB 1483 (Project-ID 442419336, EmpkinS).}
}

\author{
    \IEEEauthorblockN{Azhar Hussian, Marina Ritthaler, André Kaup, Vasileios Belagiannis}
    \IEEEauthorblockA{\textit{Friedrich-Alexander-Universität Erlangen-Nürnberg} \\
    Erlangen, Germany \\
    \{azhar.hussian, marina.ritthaler, andre.kaup, vasileios.belagiannis\}@fau.de}
}

\maketitle

\begin{abstract}
We present an image-conditioned point cloud completion approach that treats images as the primary geometric source rather than a secondary guide. To this end, we introduce an Image-to-Point (I2P) module that can reconstruct complete point clouds directly from a single RGB image, with no need for 3D inputs. Additionally, we introduce a transformer-based Point-to-Point (P2P) refinement module that uses self- and cross-attention between point tokens and image features to iteratively refine the coarse I2P output. The I2P module enables the image encoder to learn rich geometric representations, while the P2P module progressively recovers fine-grained details. Unlike existing multimodal methods that rely on auxiliary losses or fusion modules, our explicit I2P task provides a strong, geometry-aware prior based on images alone. Extensive experiments on ShapeNet-ViPC demonstrate state-of-the-art completion performance with a 12.3\% relative Chamfer Distance improvement over prior methods. Code is available at: ~\href{code}{https://github.com/AzharSindhi/I2PRef.git}
\end{abstract}


\section{Introduction}
\label{sec:intro}

Point clouds serve as a fundamental 3D representation for numerous applications, including autonomous navigation, robotic planning, augmented reality, and medical imaging. However, real-world 3D acquisition often yields incomplete scans due to occlusions, sensor noise, or limited viewpoints. Point cloud completion aims to recover the missing geometry and reconstruct a complete and high-fidelity 3D shape from a single partial observation.

\begin{figure}[h]
    \centering
    \includegraphics[width=\linewidth]{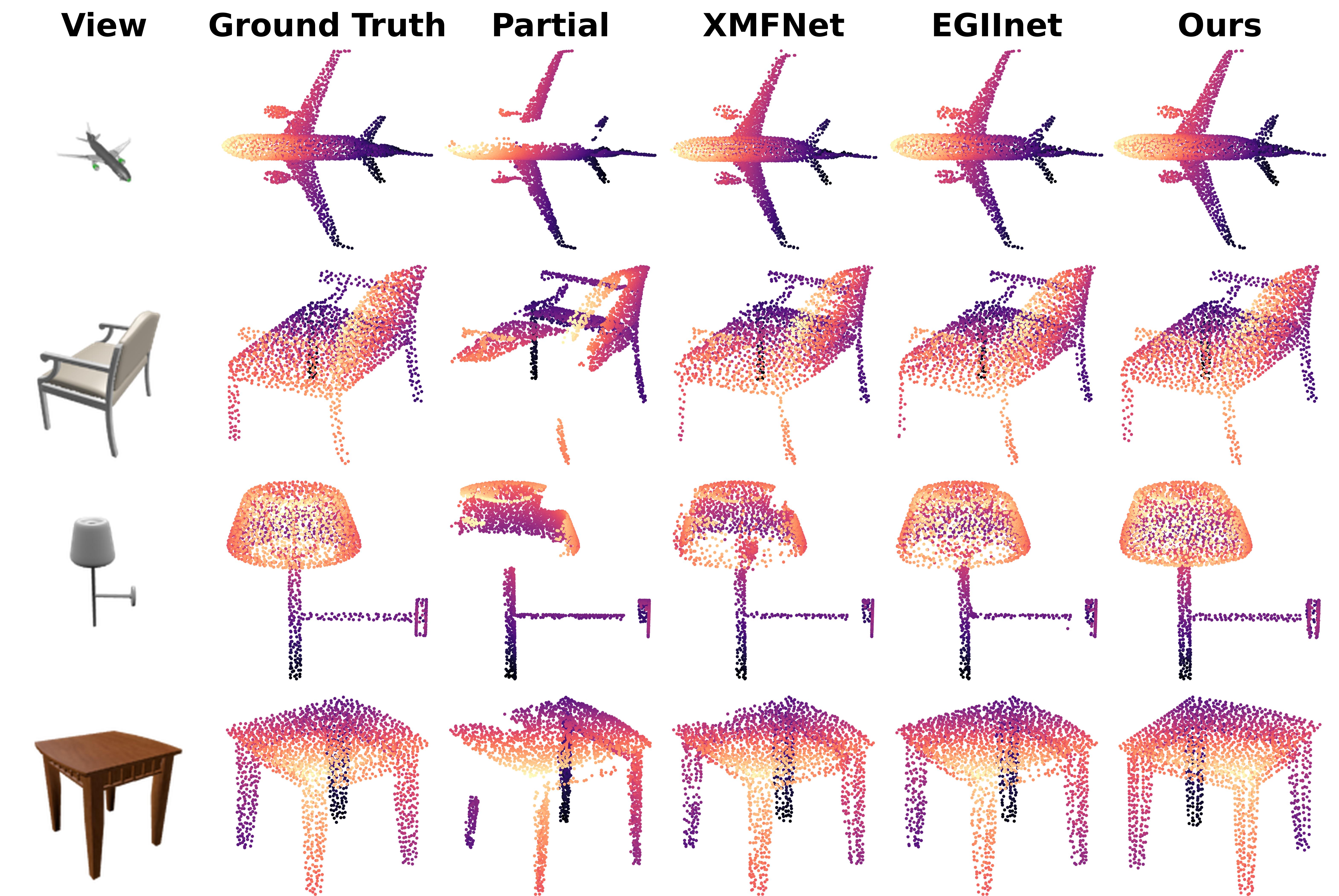}
    \caption{Our method reconstructs fine geometric details more completely and uniformly than recent approaches XMFNet~\cite{aiello2022cross} and EGIInet~\cite{xu2024}. Best viewed when zoomed.}
    \label{fig:intro}
\end{figure}

Point cloud completion has seen many progress from methods ranging from encoder-decoder based architectures~\cite{yuan2018pcn,yang2017foldingnet,Tchapmi_2019_CVPR,Huang_2020_CVPR}, coarse-to-fine refinement~\cite{Wang_2020_CVPR,Xiang2021SnowflakeNetPC}, transformer based~\cite{yu2021pointr,li2023proxyformer,zhu2023svdformer,Chen_2023_CVPR, engel2021point}, generative models such as GANs~\cite{Zhang2021Unsupervised3S}, VAEs~\cite{Mittal2022AutoSDFSP,Pan2021VariationalRP}, and more recently, diffusion models~\cite{zhang2024reversecomplete,Chu2023DiffCompleteDG,lyu2021conditional,Zhou20213DSG}. Alternatively, multi-modal methods incorporating a single RGB image have emerged as a promising direction~\cite{zhang2021view,zhu2023csdn,aiello2022cross,du2024cdpnet,xu2024}. By leveraging color, texture, and semantic cues from images, these models aim to address the ill-posed nature of inferring full 3D geometry from sparse or occluded point clouds. Although effective, these approaches typically treat image information as secondary guidance through auxiliary alignment losses or fusion modules, limiting their ability to fully exploit the rich geometric information encoded in images. In parallel, recent advances in 3D reconstruction from single or multi-view images have demonstrated remarkable success in generating high-fidelity geometry directly from 2D images~\cite{Wang2025VGGTVG}. These techniques provide compelling evidence that images alone contain sufficient geometric cues for complete 3D shape recovery, motivating their adaptation to the point cloud completion task.

Motivated by recent advances in single-image 3D reconstruction, we propose an explicit image-to-point (I2P) mechanism that treats the image as the primary source for completion. Specifically, we design a lightweight I2P module that reconstructs a complete point cloud directly from a single image, without relying on any 3D inputs or fusion. This module enables the image encoder to learn geometry-aware representations that provide a strong coarse completion serving as a prior for downstream refinement. To further enhance the reconstruction fidelity, we introduce an iterative refinement Point-to-Point (P2P) module that operates directly on the coarse I2P prediction and image features. This module refines the initial geometry through a stack of transformer blocks with self- and cross-attention, progressively correcting point positions and recovering fine-grained details while retaining a conceptually simple and modular design. We evaluate our method on the standard ShapeNet-ViPC~\cite{zhang2021view} benchmark. Extensive experiments demonstrate that our approach achieves state-of-the-art completion performance, outperforming related works in both quantitative metrics and visual quality. 

\section{Related Work}

Prior methods for point cloud completion have introduced diverse architectures to infer missing regions from partial 3D observations, pioneered by PCN~\cite{yuan2018pcn} and followed by numerous studies~\cite{zhang2024tnt,liu2020morphing,lyu2021conditional,Xiang_2021_ICCV,pan2020ecg,zhu2023svdformer,zou2025hierarchical,xiang2021snowflakenet,zhang2024tnt,li2023proxyformer}. Most approaches employ coarse-to-fine refinement strategies~\cite{liu2020morphing,lyu2021conditional,Xiang_2021_ICCV,zou2025hierarchical}, where an initial coarse shape is generated and progressively refined. Specialized methods like SnowflakeNet~\cite{xiang2021snowflakenet} introduce point deconvolution operations that model completion as a hierarchical splitting process. Following the success of transformers~\cite{Vaswani2017AttentionIA}, transformer-based architectures have gained traction, starting with PoinTr~\cite{yu2021pointr} and continuing with recent works~\cite{yu2023adapointr,wen2022pmp,zhu2023svdformer,zhang2024tnt}. Generative approaches including GANs~\cite{Zhang2021Unsupervised3S}, VAEs~\cite{Mittal2022AutoSDFSP,Pan2021VariationalRP}, and diffusion models~\cite{zhang2024reversecomplete,Chu2023DiffCompleteDG,lyu2021conditional} have also been explored. Despite the major advances in generative models, we empirically observed that diffusion models lack generalization for our task~\cite{asthana2026detecting}.

While effective, these 3D-only methods struggle to capture global object structure from highly incomplete inputs. Recent works have thus incorporated view-based image cues~\cite{zhang2021view,zhu2023csdn,aiello2022cross,xu2024} to complement partial 3D geometry. ViPC~\cite{zhang2021view} pioneered single-view image integration through cross-modality fusion, followed by methods focusing on efficient feature alignment~\cite{zhu2023csdn,aiello2022cross,xu2024}. CSDN~\cite{zhu2023csdn} introduced shape fusion and dual-refinement modules, while XMFNet~\cite{aiello2022cross} performs cross-modal interaction in a shared latent space. EGIINet~\cite{xu2024}, the most recent approach, uses unified encoding and guided information interaction to better align modalities.

Unlike prior view-based methods that treat images as auxiliary inputs to 3D encoders, our approach fully leverages the image modality by generating an initial coarse completion directly from a single RGB image. Instead of adopting existing image-to-point cloud (I2P) methods~\cite{lee2025rgb2point,chu2018generative}, we design a custom lightweight I2P network with only 6M parameters that produces a strong geometric prior from images alone. This coarse completion then guides a transformer-based refinement network that progressively enhances geometric fidelity through self- and cross-attention, achieving superior completion performance without relying on auxiliary alignment losses or fusion modules.

\section{Method}
\label{sec:method}

\begin{figure}
\centering
\includegraphics[width=\linewidth]{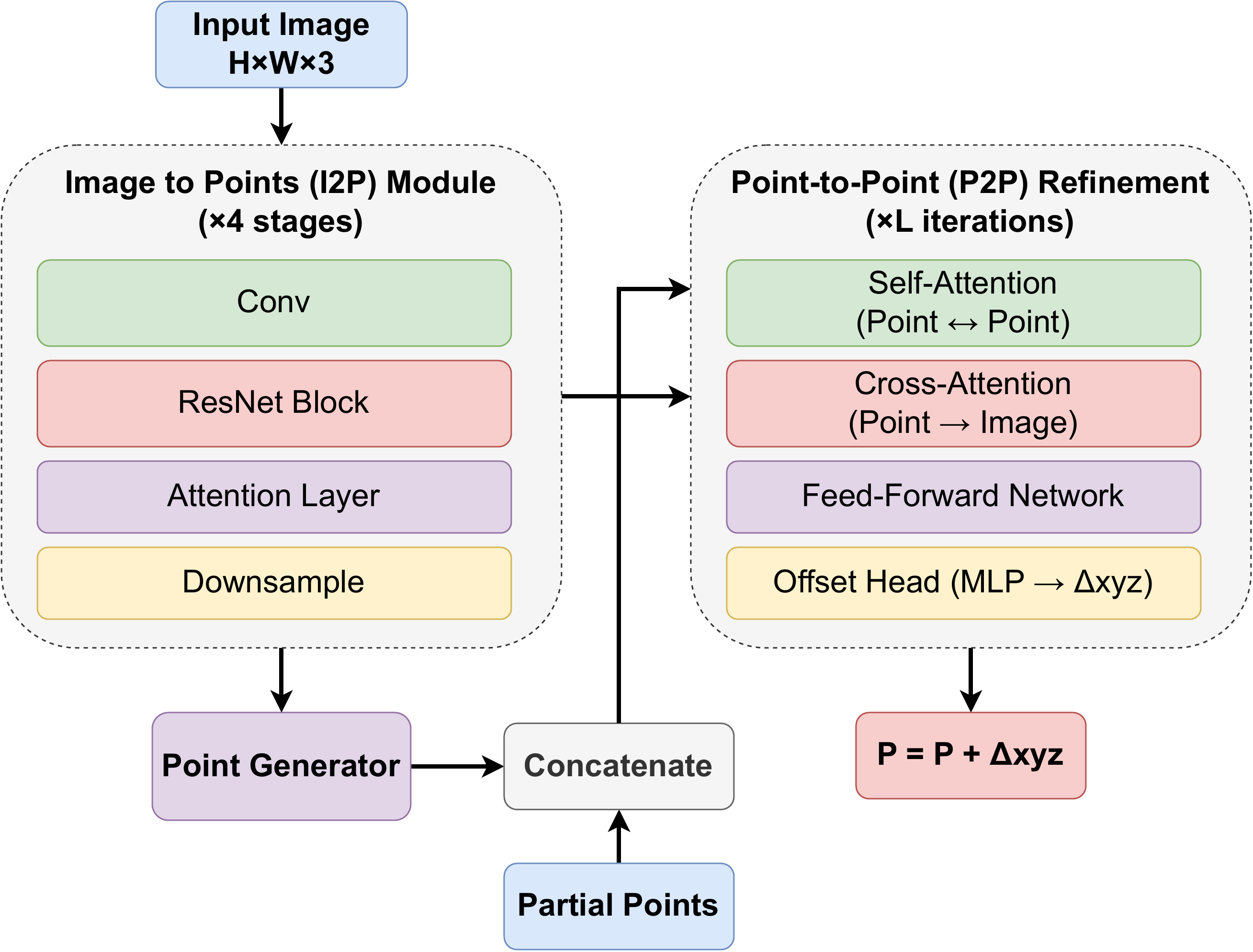}
\caption{Overview of the proposed point cloud completion framework. Given a single RGB image, we first extract hierarchical features using a U-Net style~\cite{ronneberger2015u} encoder with ResNet~\cite{He2015DeepRL} blocks and attention layers~\cite{Vaswani2017AttentionIA}. The resulting image features are passed to a point generator to produce an initial coarse point cloud. The coarse points are then iteratively refined through $L$ transformer blocks, where each block performs self-attention among point features, cross-attention to image features, and predicts coordinate offsets $\Delta xyz$ that are accumulated to refine the point positions}
\label{fig:architecture}
\end{figure}

Let $P_{i} \in \mathbb{R}^{N_{i} \times 3}$ denote the input incomplete 3D point cloud with $N_{i}$ observed points, and $I \in \mathbb{R}^{H \times W \times 3}$ be the input single RGB image providing rich geometric cues. The goal is to construct the complete point cloud $P_{c} \in \mathbb{R}^{N_{c} \times 3}$ that captures full object geometry.

Our proposed Image-to-Point (I2P) module (Sec.~\ref{sec:i2p}) reconstructs a coarse complete point cloud directly from the image $I$ alone, serving as a strong geometric prior for refinement. Furthermore, the proposed Point-to-Point (P2P) refinement network (Sec.~\ref{sec:p2p}) then iteratively improves this coarse output through stacked transformer blocks. Self-attention aggregates geometric context among points while cross-attention injects image guidance. Each iteration predicts and accumulates coordinate offsets, progressively correcting geometry to produce the final points $P_{c}$.

\subsection{Image to Points (I2P) Module}
\label{sec:i2p}

The Image-to-Points (I2P) module maps a single RGB image to a coarse 3D point cloud in a two-stage manner, decoupling image feature extraction from point generation. This design allows the image encoder to focus on learning rich spatial representations, while the point generator specializes in transforming these features into 3D coordinates.

Given an input image $\mathbf{I} \in \mathbb{R}^{H \times W \times 3}$, we employ a U-Net-style~\cite{ronneberger2015u} hierarchical encoder inspired by diffusion backbones. The encoder begins with a $7 \times 7$ convolution, followed by a sequence of downsampling stages. Each stage contains two ResNet blocks~\cite{He2015DeepRL} with RMSNorm and SiLU activations, an attention layer, and spatial downsampling. The channel dimensionality is progressively increased while spatial resolution is reduced. At the bottleneck, additional ResNet blocks with full self-attention capture global context, producing a feature map $\mathbf{F} \in \mathbb{R}^{H' \times W' \times C}$, which is then flattened into $N = H' W'$ feature vectors of dimension $C$.

To convert image features into 3D point coordinates, our framework is agnostic to the specific point generation strategy: any differentiable point generator conditioned on image features can be used. In this work, we adopt a point generator architecture similar to XMFNet~\cite{aiello2022cross}, which aggregates the encoded image features into a global latent representation and decodes it into multiple point generation branches. The generated points are concatenated with the input partial point cloud to form the coarse point cloud $\mathbf{P}_0 \in \mathbb{R}^{N \times 3}$, which serves as the input to the subsequent P2P refinement network.

\subsection{Iterative Point-to-Point (P2P) Refinement Network}
\label{sec:p2p}

The P2P network refines the coarse point cloud $\mathbf{P}_0 \in \mathbb{R}^{N \times 3}$ produced by the I2P module into a high-fidelity reconstruction $\mathbf{P}_{c} \in \mathbb{R}^{N \times 3}$. The coarse completion captures the global object structure but often leaves missing regions and oversmoothed details, especially in occluded areas. To address this, we iteratively update point positions using transformer-based refinement driven by both point features and image features.

We first map the coarse point coordinates to an embedding space via a linear projection
\begin{equation}
    \mathbf{X}_0 = \phi_{\text{emb}}(\mathbf{P}_0) \in \mathbb{R}^{N \times D},
\end{equation}
where $\mathbf{X}_0$ serves as the initial per-point embedding. At each refinement stage $l$, the current embeddings $\mathbf{X}_l$ are updated with a transformer block that applies self-attention over points and cross-attention with image features, followed by a feed-forward network and a residual connection. The self-attention uses point embeddings to form queries, keys, and values, while in the cross-attention the point embeddings act as queries and the image features $\mathbf{F}_I$ provide keys and values, enabling each point to integrate both geometric context and image guidance.

From the updated embeddings, we predict coordinate offsets and update the points:
\begin{equation}
    \Delta \mathbf{P}^{(l)} = \psi_{\text{off},l}(\mathbf{X}_l) \in \mathbb{R}^{N \times 3}, 
    \quad
    \mathbf{P}^{(l+1)} = \mathbf{P}^{(l)} + \Delta \mathbf{P}^{(l)},
\end{equation}
with $\mathbf{P}^{(0)} = \mathbf{P}_0$ and $\psi_{\text{off},l}$ a lightweight MLP mapping embeddings to 3D offsets. After $L$ refinement steps, we obtain the final refined point cloud $\mathbf{P}_{c} = \mathbf{P}^{(L)}$, which exhibits improved completeness and sharper local details.

\subsection{Loss Function}
\label{sec:loss}

The network is trained end-to-end using a composite objective that combines reconstruction supervision with uncertainty regularization. The primary Chamfer distance loss jointly supervises both the coarse I2P output ${\mathbf{P}}_0$ and the final refined reconstruction $\mathbf{P}_c$:
\begin{equation}
    \mathcal{L}_{\text{cd}} = CD(\mathbf{P}_c, \mathbf{P}_{gt}) + \alpha \, CD({\mathbf{P}}_0, \mathbf{P}_{gt}),
\end{equation}
where the symmetric Chamfer distance is defined as:
\begin{equation}
\begin{split}
    CD(\mathbf{P}_1, \mathbf{P}_2) ={}& \frac{1}{|\mathbf{P}_1|} \sum_{x \in \mathbf{P}_1} \min_{y \in \mathbf{P}_2} \|x - y\|_2^2 \\
    &+ \frac{1}{|\mathbf{P}_2|} \sum_{y \in \mathbf{P}_2} \min_{x \in \mathbf{P}_1} \|x - y\|_2^2.
\end{split}
    \label{eq:chamfer_distance}
\end{equation}

\section{Experiments}
\label{sec:experiments}

We conduct extensive experiments to evaluate our proposed method on the ShapeNet-ViPC benchmark~\cite{zhang2021view}. Our study covers quantitative and qualitative comparisons to state-of-the-art methods, evaluation of generalization to unseen object categories, and ablations of the main architectural components.

\subsection{Dataset and Setup}
We evaluate on the ShapeNet-ViPC dataset~\cite{zhang2021view}, which contains 38,328 objects across 13 categories. Following prior work~\cite{zhang2021view, aiello2022cross, xu2024}, we train on a subset of 8 categories, using 80\% of instances for training and 20\% for testing, and report both per-category and average performance under the standard protocol~\cite{zhang2021view}. Each incomplete input and its corresponding ground-truth point cloud contain $N_i = 2048$ points. The I2P module predicts $N_g = 1024$ new points, which are concatenated with 1024 points uniformly downsampled from the partial input to form a 2048-point coarse completion fed into the P2P module. The P2P module then applies $L=4$ refinement iterations. For the image input, we follow the dataset protocol and randomly select a single rendered view of resolution $224 \times 224$ from the 24 available views per object. The reconstruction loss jointly supervises the coarse I2P output and the final refined prediction using a weight $\alpha$ that is linearly annealed from 0.7 to 0.1 over training epochs, shifting emphasis from coarse structure to fine refinement. All models are trained with a batch size of 16 on 4 NVIDIA A100 GPUs.

\subsection{Evaluation Metrics}
Following prior point cloud completion work~\cite{zhang2021view,aiello2022cross,xu2024}, we evaluate performance using the Chamfer Distance (CD), as defined in Eq.~\ref{eq:chamfer_distance}, and the F1-score. The F1-score is defined as the harmonic mean of precision $P(\tau)$ and recall $R(\tau)$ at a distance threshold of $\tau = 0.001$:
\begin{equation}
\text{F1}(\tau) = \frac{2 \cdot P(\tau) \cdot R(\tau)}{P(\tau) + R(\tau)},
\end{equation}
where $P(\tau)$ denotes the mean distance from predicted points to their nearest ground truth neighbor within threshold $\tau$, quantifying the fidelity of the prediction, and $R(\tau)$ denotes the mean distance from ground truth points to their nearest predicted neighbor within threshold $\tau$, quantifying the coverage of the ground truth by the prediction. We calculate distances using the same approach as the Chamfer Distance in Eq.~\ref{eq:chamfer_distance}.

\subsection{Quantitative Results}

\begin{figure*}[!h]
    \centering
    \includegraphics[width=\linewidth]{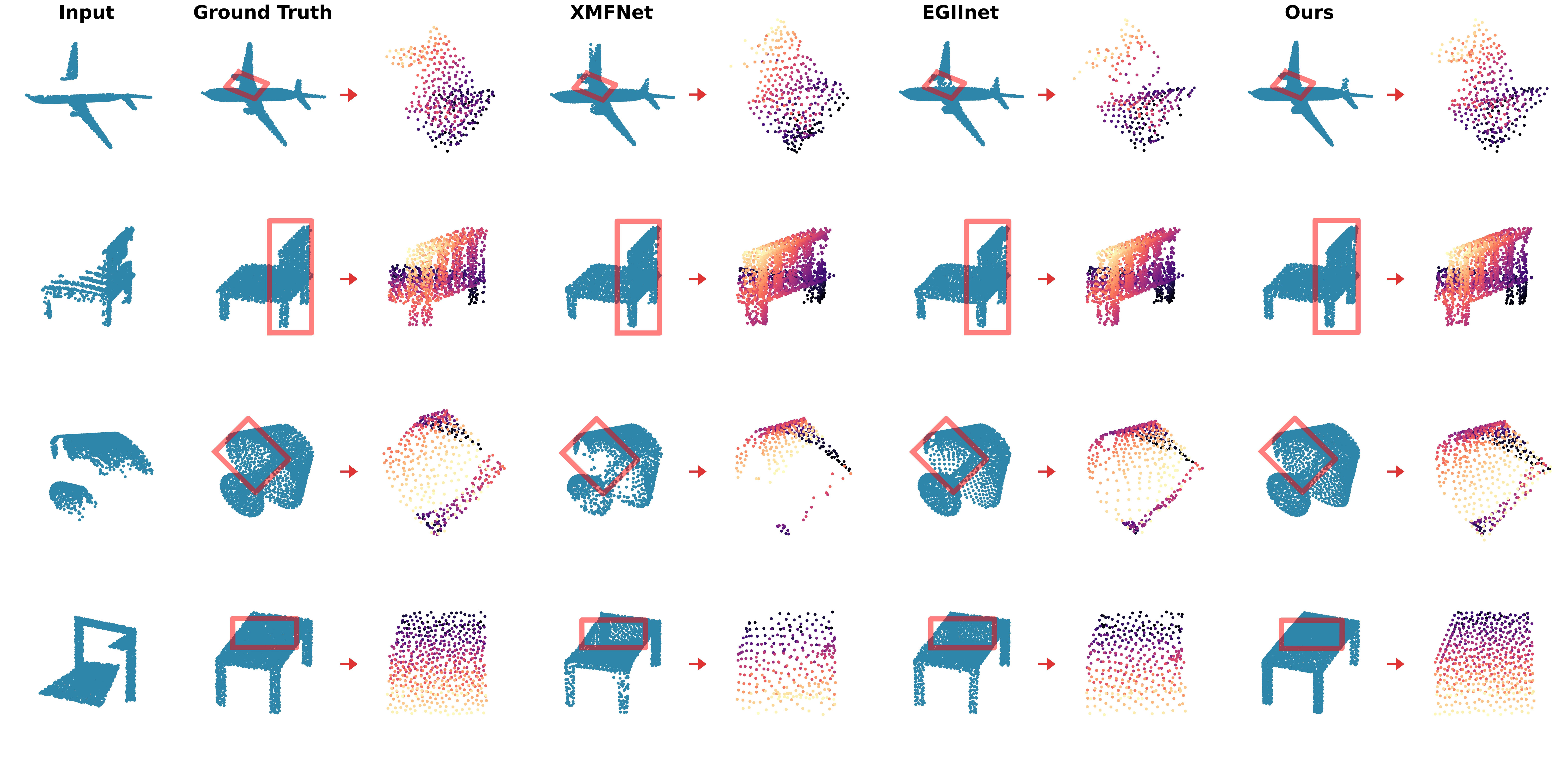}
    \caption{Qualitative comparison on four ShapeNet categories. Each row shows the partial input, ground truth, and completions from XMFNet~\cite{aiello2022cross}, EGIInet~\cite{xu2024}, and our proposed method, where red rectangles highlight distinguishing regions with their zoomed details displayed on the right. As illustrated, our method generates more complete structures with finer geometric details.}
    \label{fig:vis_zoom}
\end{figure*}

\begin{table*}[t]
  \begin{center}
  \setlength{\tabcolsep}{3.5pt}{\footnotesize
  \caption{Mean CD per point ($\text{CD}\times 10^3\downarrow$) / Mean F-Score @ 0.001$\uparrow$ on ShapeNet-ViPC.}
  \label{table:main_cd_f1}%
    \begin{tabular}{l|c c c c c c c c c}
    \toprule
    Methods&Avg&Plane&Cabinet&Car&Chair&Lamp&Sofa&Table&Watercraft \\
    \midrule
    ViPC\cite{zhang2021view}
    &3.308/0.591&1.760/0.803&4.558/0.451&3.183/0.512&2.476/0.529&2.867/0.706&4.481/0.434&4.990/0.594&2.197/0.730 \\
    CSDN\cite{zhu2023csdn}
    &2.570/0.695&1.251/0.862&3.670/0.548&2.977/0.560&2.835/0.669&2.554/0.761&3.240/0.557&2.575/0.729&1.742/0.782 \\
    XMFnet\cite{aiello2022cross}
    &1.443/0.796&0.572/0.961&1.980/0.662&1.754/0.691&1.403/0.809&1.810/0.792&1.702/0.723&1.386/0.830&0.945/0.901 \\
    EGIInet\cite{xu2024}
    &1.211/0.836&0.534/0.969&1.921/0.693&1.655/0.723&1.204/0.847&0.776/0.919&1.552/0.756&1.227/0.857&0.802/0.927 \\
    \midrule
    Ours
    &\textbf{1.062}/\textbf{0.861}&\textbf{0.474}/\textbf{0.981}&\textbf{1.620}/\textbf{0.731}&\textbf{1.510}/\textbf{0.753}&\textbf{1.136}/\textbf{0.859}&\textbf{0.597}/\textbf{0.945}&\textbf{1.421}/\textbf{0.782}&\textbf{1.086}/\textbf{0.880}&\textbf{0.650}/\textbf{0.954} \\
    \bottomrule
    \end{tabular}
}
  \end{center}  
\end{table*}

We compare our method against recent view-guided completion approaches.
Table~\ref{table:main_cd_f1} reports CD and F1-Score. As can be seen, our method consistently outperforms all competing methods across all evaluated categories and on average for both metrics, achieving a relative average improvement of \textbf{12.3\%} in CD over the most recent state-of-the-art method~\cite{xu2024}.

\vspace{0.1cm}
\noindent\textbf{Generalization to unseen categories.}  
To assess zero-shot generalization, we test the trained model on novel categories that are excluded from training. As shown in Table~\ref{table:unseen}, our method attains better performance than prior approaches, indicating better transferability to unseen object classes.

\vspace{0.1cm}
\noindent\textbf{Ablations on key components.}  
We ablate each main component using a subset of 800 training batches over 200 epochs and report average performance across all training instances. Table~\ref{table:ablations} shows that removing the I2P module, P2P refinement, or image-to-point reconstruction loss all degrade performance, with I2P supervision providing the largest contribution to coarse alignment and downstream refinement success.

\begin{table}[t]
  \begin{center}
  \setlength{\tabcolsep}{1.3pt}{\footnotesize
  \caption{Results on unknown categories measured by Chamfer Distance ($\text{CD}\times 10^{-3}\downarrow$) / (F-score @ 0.001$\uparrow$).}
  \label{table:unseen}
  \begin{tabular}{l|c c c c c}
    \toprule
    Methods&Avg&Bench&Monitor&Speaker&Cellphone \\
    \midrule
    ViPC\cite{zhang2021view} & 4.601/0.498 & 3.091/0.654 & 4.419/0.491 & 7.674/0.313 & 3.219/0.535 \\
    CSDN\cite{zhu2023csdn} & 3.656/0.631 & 1.834/0.798 & 4.115/0.598 & 5.690/0.485 & 2.985/0.644 \\
    XMFnet\cite{aiello2022cross} & 2.671/0.710 & 1.278/0.862 & 2.806/0.677 & 4.823/0.556 & 1.779/0.748 \\
    EGIInet\cite{xu2024} & 2.354/0.750 & 1.047/0.902 & 2.513/0.716 & 4.282/0.591 & 1.575/0.792 \\
    \midrule
    Ours & \textbf{2.287}/\textbf{0.760} & \textbf{0.987}/\textbf{0.910} & \textbf{2.361}/\textbf{0.732} & \textbf{4.214}/\textbf{0.605} & 
    \textbf{1.570}/\textbf{0.801} \\
    \bottomrule
    \end{tabular}
    }
  \end{center}
\end{table}

\begin{table}
    \centering
    \caption{Ablation study results ($\text{CD}\times 10^3\downarrow$).}
    \label{table:ablations}

    \begin{tabular}{l|c}
    \hline
    Methods & Avg \\
    \hline
    
    Ours 
        & \textbf{1.644} \\

    Ours w/o recon
        & 1.758 \\
        
    I2P only
        & 2.551 \\

    P2P only
        & 3.871 \\
    \hline
    \end{tabular}
\end{table}


Furthermore, Figure~\ref{fig:vis_zoom} shows qualitative comparisons on four ShapeNet categories. Our method produces more complete and geometrically accurate reconstructions compared to XMFNet~\cite{aiello2022cross} and EGIInet~\cite{xu2024}, particularly in fine-grained structural details.

\section{Conclusion}

In this paper, we presented an image-conditioned point cloud completion approach that treats images as the primary geometric source rather than a secondary guide. We introduced an Image-to-Point (I2P) module that reconstructs complete point clouds directly from a single RGB image, with no need for 3D inputs. Additionally, we introduced a transformer-based Point-to-Point (P2P) refinement module that uses self- and cross-attention between point tokens and image features to iteratively refine the coarse I2P output. 
Extensive experiments on ShapeNet-ViPC~\cite{zhang2021view} demonstrated state-of-the-art performance with a 12.3\% relative Chamfer Distance improvement over prior methods, alongside superior visual completeness and detail preservation. Our approach demonstrates strong generalization to unseen categories in ShapeNet-ViPC and effectively handles challenging regions where existing approaches struggle. Future work will explore multi-view extensions and real-world LiDAR integration for practical applications.



{
    \small
    \bibliography{main}
}

\end{document}